\documentclass[sigconf]{acmart}
\AtBeginDocument{%
  \providecommand\BibTeX{{%
    \normalfont B\kern-0.5em{\scshape i\kern-0.25em b}\kern-0.8em\TeX}}}

\setcopyright{acmcopyright}
\copyrightyear{2022}
\acmYear{2022}
\setcopyright{acmcopyright}\acmConference[MM '22]{Proceedings of the 30th ACM International Conference on Multimedia}{October 10--14, 2022}{Lisboa, Portugal}
\acmBooktitle{Proceedings of the 30th ACM International Conference on Multimedia (MM '22), October 10--14, 2022, Lisboa, Portugal}
\acmPrice{15.00}
\acmDOI{10.1145/3503161.3548281}
\acmISBN{978-1-4503-9203-7/22/10}
\acmSubmissionID{2293}

\usepackage{caption}
\usepackage{subcaption}
\usepackage{multirow}
\usepackage{multicol}
\usepackage{enumitem}
\usepackage{float}
\usepackage{xcolor,colortbl}
\usepackage{scalerel}
\begin{document}

\title{Target-Driven Structured Transformer Planner for Vision-Language Navigation}

%
\author{Yusheng Zhao}
\authornote{Equal contribution.}
\affiliation{%
  \institution{Institute of Artificial Intelligence,}
  \institution{Hangzhou Innovation Institute, Beihang University}
  \city{Beijing}
  \country{China}
}
\email{zhaoyusheng@buaa.edu.cn}

\author{Jinyu Chen}
\authornotemark[1]
\affiliation{%
  \institution{Institute of Artificial Intelligence,}
  \institution{Hangzhou Innovation Institute, Beihang University}
  \city{Beijing}
  \country{China}
}
\email{chenjinyu@buaa.edu.cn}

\author{Chen Gao}
\affiliation{%
  \institution{Institute of Artificial Intelligence,}
  \institution{Hangzhou Innovation Institute, Beihang University}
  \city{Beijing}
  \country{China}
}
\email{gaochen.ai@gmail.com}

\author{Wenguan Wang}
\authornote{Corresponding author.}
\affiliation{%
  \institution{ReLER, AAII}
  \institution{University of Technology Sydney}
  \city{Sydney}
  \country{Australia}}
\email{wenguanwang.ai@gmail.com}

\author{Lirong Yang}
\affiliation{%
  \institution{Meituan Inc.}
  \city{Beijing}
  \country{China}}
\email{yanglirong@meituan.com}

\author{Haibing Ren}
\affiliation{%
  \institution{Meituan Inc.}
  \city{Beijing}
  \country{China}}
\email{renhaibing@meituan.com}

\author{Huaxia Xia}
\affiliation{%
  \institution{Meituan Inc.}
  \city{Beijing}
  \country{China}}
\email{xiahuaxia@meituan.com}

\author{Si Liu}
\affiliation{%
  \institution{Institute of Artificial Intelligence,}
  \institution{State Key Laboratory of Virtual Reality Technology and Systems, SCSE, Beihang University}
  \city{Beijing}
  \country{China}
}
\email{liusi@buaa.edu.cn}

%
\renewcommand{\shortauthors}{Y. Zhao, J. Chen, C. Gao, W. Wang, L. Yang, H. Ren, H. Xia, S. Liu}

\begin{abstract}
Vision-language navigation is the task of directing an embodied agent to navigate in 3D scenes with natural language instructions. For the agent, inferring the long-term navigation target from visual-linguistic clues is crucial for reliable path planning, which, however,  has rarely been studied before in literature.
In this article, we propose a Target-Driven Structured Transformer Planner (TD-STP) for long-horizon goal-guided and room layout-aware navigation. Specifically, we devise an Imaginary Scene Tokenization mechanism for explicit estimation of the long-term target (even located in unexplored environments). In addition, we design a Structured Transformer Planner which elegantly incorporates the explored room layout into a neural attention architecture for structured and global planning. Experimental results demonstrate that our TD-STP substantially improves previous best methods' success rate by \textbf{$2$\%} and \textbf{$5$\%} on the test set of R2R and REVERIE benchmarks, respectively. Our code is available at \url{https://github.com/YushengZhao/TD-STP}.
\end{abstract}

%




\maketitle

\begin{figure*}
  \centering
  \includegraphics[width=0.9\linewidth]{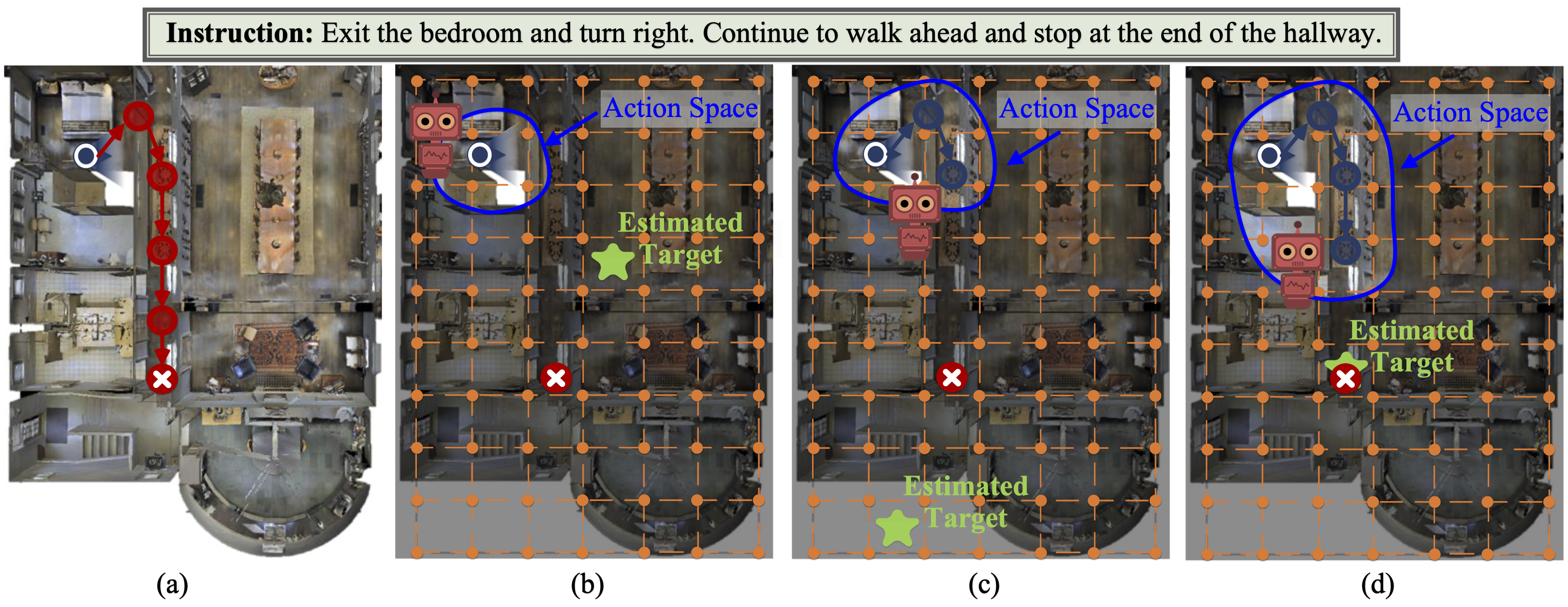}
  \vspace{-2mm}
        \caption{(a) Ground truth navigation trajectory. (b) At the beginning of navigation, our agent gives a rough estimation of the final destination, which serves as guidance for long-term planning. (c)-(d) With the navigation progresses, our agent can gradually refine the target estimation, so as to provide more and more precise guidance for navigation planning.}
  \label{fig:teaser}
  \vspace{-2mm}
\end{figure*}

\section{Introduction}
Recent years have witnessed increasing interest in 
the creation of an embodied agent which learns to actively solve various challenging tasks within its environment. A number of simulators~\cite{chang2017matterport3d, kolve2017ai2thor, Savva_2019_habitat} and datasets~\cite{xia2018gibson, dai2017scannet} have been proposed, backing such tasks as navigation~\cite{anderson2018vln, gupta2017cognitive}, multi-agent cooperation~\cite{wang2021collaborative, busoniu2008comprehensive, mordatch2018emergence}, interactive learning~\cite{deitke2020robothor},
and visual grounding~\cite{shridhar2020alfred, achlioptas2020referit3d, he2021transrefer3d, luo20223d}.

Vision-Language Navigation (VLN), one of the most representative embodied AI tasks, poses particular challenges as it requires the agent to navigate visual environments by following linguistic instructions. Current prevalent VLN agents~\cite{chen2021hamt, Hong2021vlnbert, Hao_2020_prevalent} are built upon a cross-modal transformer architecture which makes only use of language instructions and historical perception for decision making. Though effective in cross-modality alignment and reasoning over past observation, they lack the sense of long-horizon goal. Take Figure~\ref{fig:teaser} (a) as an example. Given the instruction ``Exit the bedroom and turn right. Continue to walk ahead and stop at the end of the hallway'', human beings can imagine a likely long-term target even before starting navigation. At the very beginning, the estimation of the final destination may not be perfect. However, it can be progressively corrected with the navigation proceeded, and, undoubtedly, serves as the basis of our planning ahead. Inspired by this, in this work, we aim to equip the VLN agent with the ability of "imaging the long-horizon future". 

Achieving long-term target-driven planning is not easy. First, estimating the long-term targets is challenging. The final destination is located at the unexplored environments; the candidate positions are countless. Second, how to make full use of the estimated long-term target to assist navigation is also an open question; prevalent methods are typically aware of \textit{present} (\emph{i.e.}, current observation) and \textit{past} (\emph{i.e.}, navigation history), yet paying less attention to the \textit{future}. To tackle these challenges, we propose an Imaginary Scene Tokenization (IST) mechanism which enables long-term target representation and prediction, as well as accommodates target-driven planning within the prevalent Transformer-based navigation framework. IST discretizes the unexplored area into a fixed-size grid. Each grid cell is represented by a target token that captures the imagined layout of the cell. The target tokens are fed into a cross-modal transformer along with other visual-linguistic cues to model the \emph{history}, \emph{present} and \emph{future} of the navigation. Then these tokens are used to estimate whether the navigation target is in its cell, so as to enable global planning. 

In addition, as the VLN agent faces structured environments, understanding the topology of the environment is crucial for the success of navigation. However, existing methods either arrange the historical observations in a sequential manner~\cite{devlin2018bert, moudgil2021soat,lin2021multimodal}, or adopt complicated modules (\emph{e.g.}, graph neural networks) for modeling environment layouts~\cite{wang2021structured, hong2020language,chen2022think}. Differently, we develop a Structured Transformer Planner (STP), where the position-embedded visual observations are used as input tokens, and the geometric relations (local connectivity) among the navigation locations are elegantly formulated as the directional attention among input tokens. With such a design, the agent is able to not only gain a comprehensive understanding of the environment layout, but also easily revisit the past visited locations (see Figure~\ref{fig:teaser} (b)-(d)). 

The integration of STP and  IST leads to a Target-Driven Structured Transformer Planner (TD-STP), which allows for long-horizon goal-guided and environment layout-aware navigation. Experiments on Room-to-Room (R2R)~\cite{anderson2018vln} and REVERIE~\cite{qi2020reverie} datasets show that TD-STP achieves state-of-the-art performance on the test sets.  Our code is available at \url{https://github.com/YushengZhao/TD-STP}.

    
    

\begin{figure*}[ht]
    \includegraphics[width=0.95\linewidth]{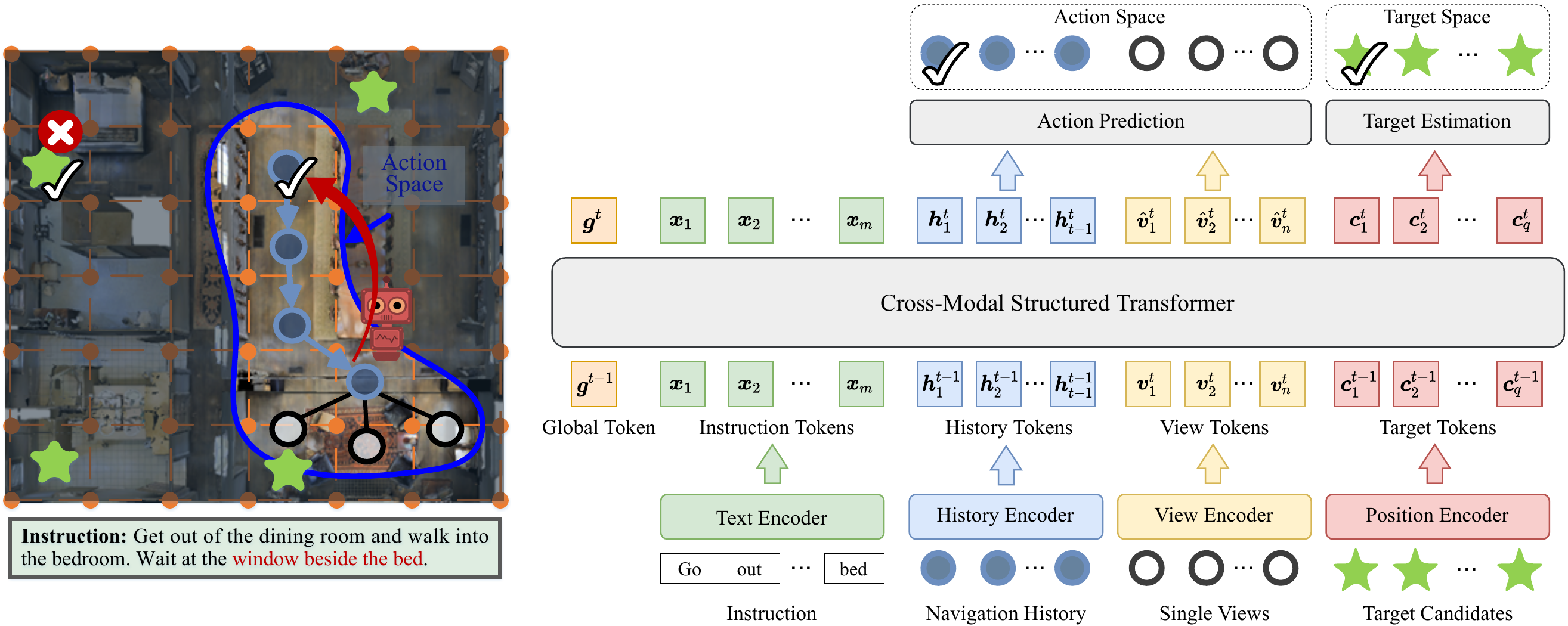}
    \caption{An overview of our TD-STP model. At time step $t$, five types of tokens (\emph{i.e.}, the global token, instruction tokens, target tokens, history tokens, and view tokens) are sent into the cross-modal transformer, simultaneously, to predict an action decision. Note that the action space contains both historical and current visible viewpoints, and the target estimation considers all the linguistic and visual observations to provide global guidance for long-term planning.}
    \label{fig:model-overview}
    \vspace{-1mm}
\end{figure*}

\section{Related Work}
The release of R2R dataset~\cite{anderson2018vln} stimulated the study of VLN. Various datasets have been later proposed to cover different navigation scenarios with high-level instruction~\cite{qi2020reverie}, multilingual instruction~\cite{ku2020room}, dialog-based instruction~\cite{thomason2020vision}, and fine-grained instruction~\cite{zhu2020babywalk}.

Meanwhile, numerous navigation agents have been successfully developed.
Some works focus on learning better representations~\cite{zhu2021diagnosing, hu2019you, wang2020environment, qi2020object}. For example, Wang \emph{et al}.~\cite{wang2020environment} propose to learn environment-agnostic representation and use multi-task learning to further enrich the representation. 
Some other works instead explore smarter path planning strategies~\cite{wang2018look, koh2021pathdreamer, anderson2019chasing, ke2019tactical, ma2019regretful, wang2020active, wang2021structured, krantz2021waypoint}. For instance, Ma \emph{et al}.~\cite{ma2019regretful}  employ a backtracking strategy that allows the agent to decide whether to continue moving forward or roll back to a previous state. 
To address the issue of data scarcity, several works adopt auxiliary-task learning~\cite{zhu2020auxrn, wang2019reinforced, chen2021hamt} and data augmentation~\cite{fried2018speaker, huang2019transferable, fu2020counterfactual, liu2021vision, tan2019learning, wang2022counterfactual} techniques. For example, Zhu \emph{et al}.~\cite{zhu2020auxrn} use progress estimation and angle prediction as auxiliary tasks. Fried \emph{et al}.~\cite{fried2018speaker} learn a speaker module to create instructions for unlabeled paths as extra training samples. For conducting long-term reasoning over past, structured observations, mapping based agents are built~\cite{wang2021structured, hong2020language, deng2020evolving}. For instance, Wang \emph{et al}.~\cite{wang2021structured} store past observations in an external, graph-like memory and use a graph neural network for structured reasoning. Moreover, different training strategies are also explored~\cite{anderson2018vln, wang2018look, wang2019reinforced, zhu2020babywalk, zhang2021curriculum}, including imitation learning (IL)~\cite{anderson2018vln}, reinforcement learning (RL)~\cite{wang2018look}, hybrid of IL and RL~\cite{wang2019reinforced}, and curriculum learning~\cite{zhu2020babywalk,zhang2021curriculum}. More recently, some researcher made use of extensive, unpaired image and text data for pre-training, and then fine-tune on the limited, labeled VLN data~\cite{Hao_2020_prevalent, chen2021hamt, guhur2021airbert, majumdar2020improving}, achieving promising results.  

With the success of transformer~\cite{vaswani2017attention} in computer vision~\cite{liu2021swin, dosovitskiy2020image, carion2020end}, natural language processing~\cite{devlin2018bert, radford2019language} and cross-modal tasks~\cite{lu2019vilbert}, transformer-based agents have been increasingly popular in VLN task~\cite{Hao_2020_prevalent, chen2021hamt, guhur2021airbert, moudgil2021soat, majumdar2020improving, Hong2021vlnbert}. 
A core challenge is how to incorporate the navigation history into the decision-making process under the transformer architecture. Some works equip transformer-based agents with recurrent modules. For example, Hong \emph{et al}.~\cite{Hong2021vlnbert} and Moudgil \emph{et al}.~\cite{moudgil2021soat} adopt a recurrent state that is updated at each navigation step, and treat the states at different steps as the input tokens of the transformer. Though straightforward, the recurrent state inevitably loses useful information when compressing the past history into the state vector. An alternative is to keep a full sequence of navigation history. For example, Chen \emph{et al}.~\cite{chen2021hamt} employ a hierarchical transformer to encode the full navigation history as a sequence of history tokens. Similarly, some methods adopt a memory bank to store a whole sequence of past action-observation tokens~\cite{lin2021multimodal, lin2021scene, pashevich2021episodic}. 

Our TD-STP distinguishes itself from previous models in its ability of long-term target-driven navigation planning, based on the explicit estimation of the final navigation targets and structured modeling of explored environment. Most LSTM-based methods~\cite{zhu2020auxrn, anderson2018vln, fried2018speaker, wang2020environment, wang2021structured, chen2022reinforced} and transformer-based methods~\cite{Hong2021vlnbert, chen2021hamt, lin2021multimodal, pashevich2021episodic, moudgil2021soat} focus on reasoning over past observations, lacking the ability of "imaging the future". In contrast,  our agent learns to explicitly predict the long-horizon navigation target, which allows for reasoning over past and planning ahead. This idea is powerful and principled, distinctively differentiates our approach from most existing navigation agents. We notice that some previous works~\cite{wang2021structured, hong2020language} are also aware of modeling the environment layout. Compared to these works, which often use complex graph neural networks, our model naturally incorporates environment layouts into cross-modal transformer, by using local connectivity between navigation locations to guide the information flow between input tokens.  A concurrent work~\cite{chen2022think} uses double cross-modal encoders for global action prediction and local action prediction, respectively. In contrast, we jointly model the environment topology and the whole action space in a single transformer, making our method elegant and flexible.

\section{Method}
\subsection{Problem Setup and Overview}
\textbf{Problem Setup.} In the VLN task, the agent is required to navigate to the target location according to a natural language instruction. We denote the textual embeddings of the instruction as $\boldsymbol{x}_0, \boldsymbol{x}_1, \boldsymbol{x}_2, \dots, \boldsymbol{x}_m$, where $\boldsymbol{x}_0$ is the sentence embedding and $m$ is the length of the instruction. At each time step $t$, the agent observes a panoramic view of the current location, consisting of $36$ single views, among which the first $k^t$ are navigable. We denote the features of these single views as ${\boldsymbol{v}}_1^t, {\boldsymbol{v}}_2^t, \dots, {\boldsymbol{v}}_{n}^t,~n=36$. In order to focus on high-level planning, in~\cite{anderson2018vln}, the environment is assumed to be a set of discrete points and their navigability is given.

\noindent\textbf{Overview.} Figure~\ref{fig:model-overview} provides an overview of the proposed TD-STP model. TD-STP uses a cross-modal structured transformer, similar to~\cite{chen2021hamt}. At time step $t$, $5$ types of tokens are sent to the transformer, \emph{i.e.}, the global token $\boldsymbol{g}^{t-1}$, the instruction tokens $\boldsymbol{x}_1, \boldsymbol{x}_2, \dots, \boldsymbol{x}_m$, the target tokens $\boldsymbol{c}_1^{t-1}, \boldsymbol{c}_2^{t-1}, \dots, \boldsymbol{c}_q^{t-1}$ ($q$ is the number of target candidates), the history tokens $\boldsymbol{h}_1^{t-1}, \boldsymbol{h}_2^{t-1}, \dots, \boldsymbol{h}_{t-1}^{t-1}$ and the view tokens $\boldsymbol{v}_1^t, \boldsymbol{v}_2^t, \dots, \boldsymbol{v}_n^t$. Note that superscripts are used to denote the time step of a token. The instruction tokens are kept constant through time to reduce computation. 
The global token $\boldsymbol{g}^{t-1}$, history tokens $\boldsymbol{h}^{t-1}$, and target tokens $\boldsymbol{c}^{t-1}$ are the outputs of previous time step $t-1$.
View tokens $\boldsymbol{v}^t$ are newly obtained at time step $t$. The instruction tokens are the output of a BERT model~\cite{devlin2018bert}, and the global token is initialized as the sentence embedding $\boldsymbol{g}^0 = \boldsymbol{x}_0$. Three other types of tokens are discussed in the following subsections.

\subsection{Structured Transformer}
\label{sec:structured-transformer}
\begin{figure}[t]
    \centering
    \includegraphics[width=0.99\linewidth]{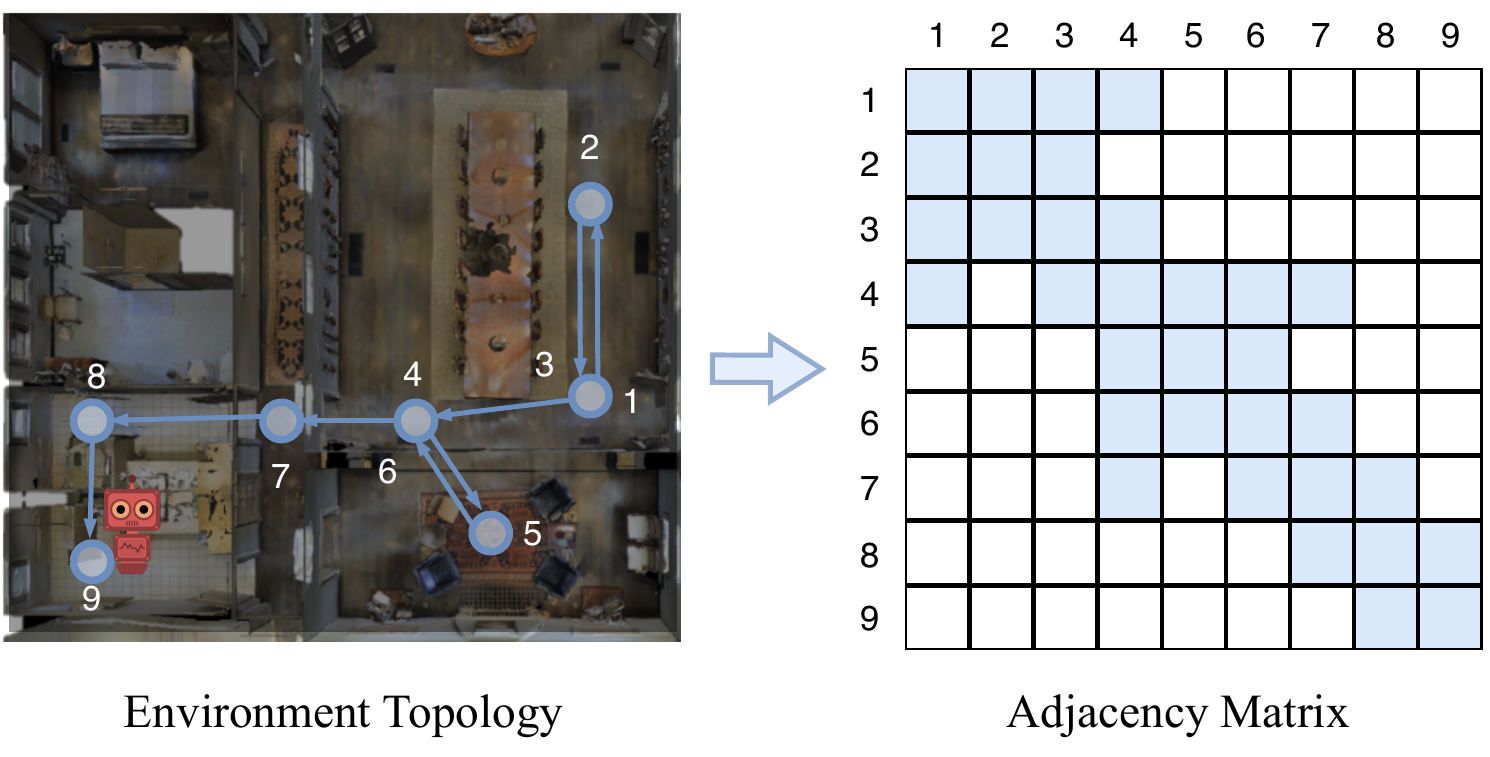}
    \vspace{-1mm}
    \caption{The environment topology (left) is embedded in the graph $\mathcal S^t$. The adjacency matrix (right) of this graph is used in the attention mask of the transformer.}
    \label{fig:topology}
    \vspace{-1mm}
\end{figure}

To capture the structured environment layouts, our TD-STP constructs and maintains a structured representation of the explored area with the transformer architecture, which is achieved via deriving a graph from navigation history and incorporating its topology into the transformer.
Concretely, at time step $t$, the model constructs a graph $\mathcal S^t$, as shown in~Figure~\ref{fig:topology}, where the nodes represent previously visited locations and the edges represent the navigability of those locations. Thus the topology of this graph can be decomposed into two parts, \emph{i.e.}, the position of each node and their adjacency.

At time step $t$, the history token $\boldsymbol{h}_t^t$ is constructed using panoramic view embeddings, action embeddings, temporal embeddings, and positional embedding as described below:
\begin{equation}
    \begin{aligned}
        \boldsymbol{h}_t^t = f_{V}(\boldsymbol{v}_1^t, \dots, \boldsymbol{v}_{n}^t) + f_A(\boldsymbol{r}^t) + f_T(t) + f_P(l^t),
    \end{aligned}
    \label{eq:hist_emb}
\end{equation}
where $f_{V}$ is a panoramic visual feature extractor (as in~\cite{chen2021hamt}), $\boldsymbol{r}^t = (\sin \theta^t, \cos \theta^t, \sin \phi^t, \cos \phi^t)$ is the moving direction ($\theta$ and $\phi$ are heading and elevation) at time step $t$, $f_A(\cdot)$ is the action encoder consisting of a linear layer and a layer normalization, and $f_T(\cdot)$ is the temporal encoder that maps the time step integer into a feature vector.
Note that an additional position encoder $f_P(\cdot)$ is utilized to incorporate the spatial location $l^t$ (\emph{e.g.} the starting position has the location of $(0,0)$; a specific node position might have the location of $(2, -9)$) information of the current node, which is relative to the starting location. The position encoder $f_P(\cdot)$ consists of a linear projection and a layer normalization.

To further incorporate the adjacency information of each navigable viewpoint into the transformer, our TD-STP leverages the attention masks to control the information flow among tokens. 
We first define the adjacency of history tokens. At time step $t$, the input history tokens of the transformer are $\boldsymbol{h}_1^{t-1}, \boldsymbol{h}_2^{t-1}, \dots, \boldsymbol{h}_{t-1}^{t-1}$, which correspond to $t-1$ historically visited locations $l_1, l_2, \dots, l_{t-1}$. The adjacency matrix of history tokens at time step $t$ is defined as a $(t-1) \times (t-1)$ matrix $\boldsymbol{E}$. If a navigation viewpoint $l_j$ is navigable from $l_i$, $\boldsymbol{E}_{ij}=1$, and otherwise $\boldsymbol{E}_{ij}=0$, as shown in Figure~\ref{fig:topology}. 

The cross-modal transformer has an attention mask matrix $\boldsymbol{M}$ that controls whether one token can attend to another in the attention layer of the transformer. Formally, if the $i$-th token of the transformer input can attend the $j$-th token, $\boldsymbol{M}_{ij} = 1$, and otherwise $\boldsymbol{M}_{ij} = 0$. Besides, the attention mask matrix $\boldsymbol{M}$ has a submatrix $\boldsymbol{M}_H$, which controls whether one history token can attend to another. Thus we incorporate the adjacency information into the transformer by masking $\boldsymbol{M}_H$ with the adjacency matrix $\boldsymbol{C}$:
\begin{equation}
    \begin{aligned}
        \boldsymbol{M}_H \leftarrow \boldsymbol{M}_H * \boldsymbol{C},
    \end{aligned}
\end{equation}
where $*$ means element-wise multiplication. In this way, two nonadjacent history tokens cannot directly affect each other during the attention computation, and the information is enforced to flow over the encoded topology. This design leads to an elegant and structured transformer-based navigator. 

\subsection{Global Action Space}
The structured transformer allows the agent to have direct access to structured information of the past, which delivers a global action space, where the agent can choose from not only adjacent locations but also previously visited locations. The global action space is flexible as it allows the agent to move off from the current direction and `jump' to previous locations, as shown in Figure~\ref{fig:model-overview}.

Most transformer-based agents proposed in previous works~\cite{chang2017matterport3d, Hong2021vlnbert, moudgil2021soat, lin2021multimodal, pashevich2021episodic} make decisions/actions from a local action space, in which the agent chooses one of the navigable single views from the current observation to walk into. For simplicity, we introduce a transformation noted as $\tau$ that maps the token to its corresponding location that is part of the action space. With this notation, the local action space at time step $t$ can be formulated as:
\begin{equation}
    \begin{aligned}
        \mathcal A_L^t = \{ \tau(\hat {\boldsymbol{v}}_1^t), \tau(\hat {\boldsymbol{v}}_2^t), \dots, \tau(\hat{\boldsymbol{v}}_{k^t}^t) \},
    \end{aligned}
\end{equation}
where $\hat{\boldsymbol{v}}_i^t$ denotes the $i$-th view token in the transformer output, and $k^t$ is the total number of navigable single views at time step $t$. 

Different from the local action space, TD-STP delivers a global action space, in which the agent has direct access to history. Mathematically, the global action space is defined as follows:
\begin{equation}
    \begin{aligned}
        \mathcal A_G^t = \{ \tau(\hat{\boldsymbol{v}}_1^t), \dots, \tau(\hat{\boldsymbol{v}}_{k^t}^t), \tau(\boldsymbol{h}_1^t), \tau(\boldsymbol{h}_2^t), \dots, \tau(\boldsymbol{h}_{t-1}^t) \}.
    \end{aligned}
\end{equation}
The global action space makes it possible for the agent to backtrack by choosing historically visited locations when it finds itself walking on the wrong path for several steps. Although a view token and a history token may correspond to the same navigation viewpoint, they are treated as two different actions since they contain different semantics. Specifically, the history token means backtracking, which indicates that the current path might be wrong, whereas the view token indicates that the current path matches the instruction. Therefore, with the action space expanded, the probability of each action is computed as:
\begin{equation}
    \begin{aligned}\label{Eq:MLP1}
    \pi (a^t;\Theta) = \text{softmax}\{MLP(\tau^{-1}(a^t) * \boldsymbol{g^t})\}, ~a^t\in \mathcal A_G^t ,
    \end{aligned}
\end{equation}
where $\pi$ is the policy function, $\Theta$ is the parameters of the model, $MLP$ is a multi-layer perceptron, $\tau^{-1}$ maps the action back to the corresponding token, and $\boldsymbol{g^t}$ is the global token.

\subsection{Imaginary Scene Tokenization Mechanism}
An important part of long-term target-driven navigation is to explicitly model the possible long-term targets, which is achieved via the proposed Imaginary Scene Tokenization (IST) mechanism. 
Specifically, the core problem is how to model the unexplored area since the exact topology and visual information of the unexplored area are unknown. 
In this subsection, we elaborate on how the IST mechanism solves the problem by \emph{discretizing}, \emph{imagining} and \emph{refining} the unexplored scene representation according to the instruction and on-the-fly collected visual clues.

\begin{figure}[t]
    \centering
    \includegraphics[width=0.95\linewidth]{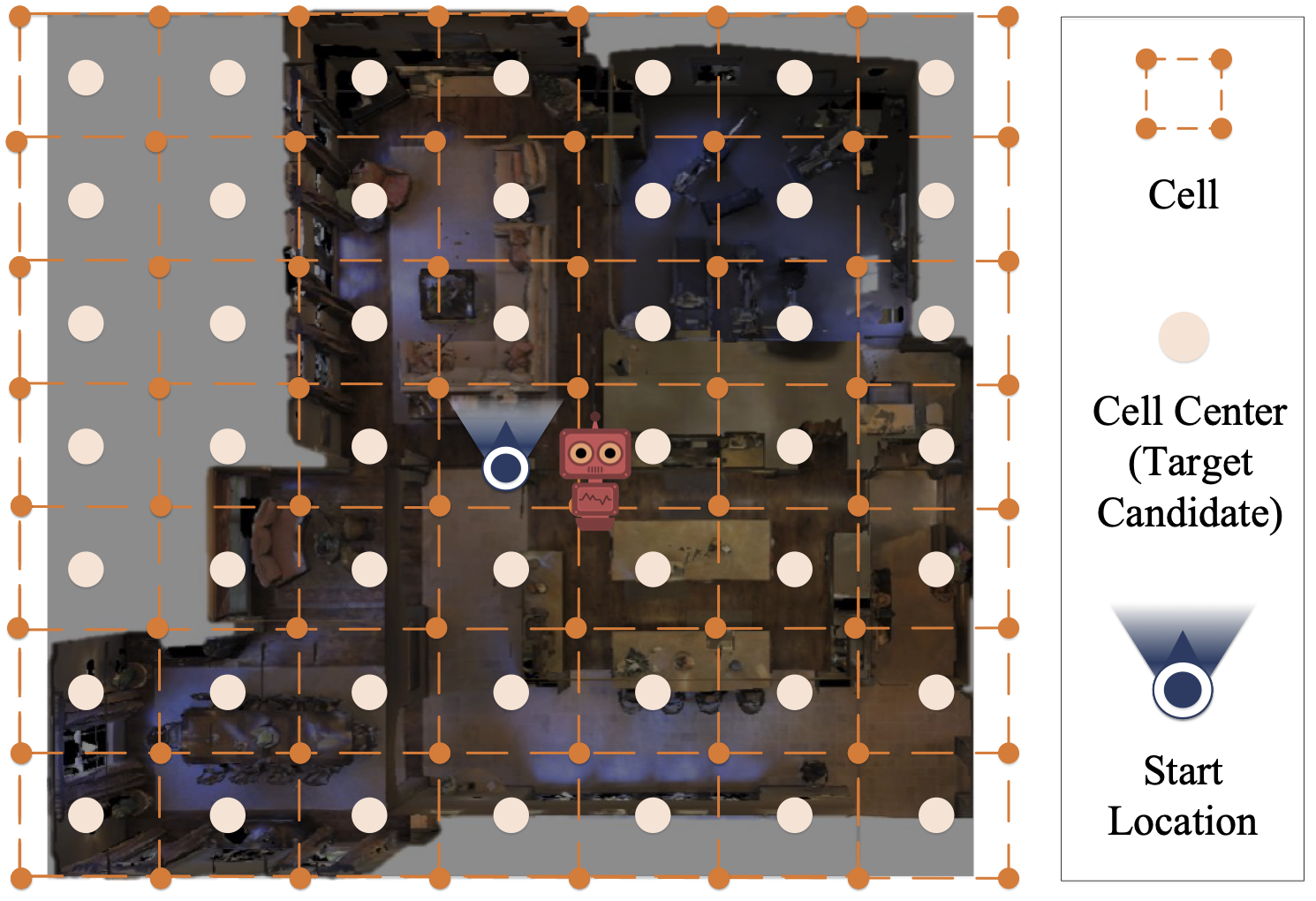}
    \vspace{-1mm}
    \caption{The agent discretizes the environment as a fixed-size $d\times d$ grid, centered at the starting point, forming $d^2$ cells. Each cell has a cell center, which is represented by a target token and treated as a long-term navigation target candidate.}
    \label{fig:ist}
    \vspace{-1mm}
\end{figure}

As shown in Figure~\ref{fig:ist}, IST first discretizes the environment into a $d\times d$ grid, which is fixed in size and covers the navigation region. The grid has $d^2$ cells and the cell centers are the possible targets of navigation. The targets are spaced $s$ meters apart, and each of them is represented by a target token. At the beginning of navigation, target tokens $\boldsymbol{c}_1^0, \dots, \boldsymbol{c}_q^0,~q=d^2$ are constructed using the positional embeddings of the targets, which is formulated as:
\begin{equation}
    \begin{aligned}
    \boldsymbol{c}_i^0 = f_P(l_i) * \boldsymbol{x}_0, \quad i\in\{1, 2, \dots, q\},
    \end{aligned}
\end{equation}
where $f_P$ is the positional encoder in Eq.~\ref{eq:hist_emb}, $l_i$ is the spatial location of $i$-th target (under the same coordinate system mentioned in Eq.~\ref{eq:hist_emb}), and $\boldsymbol{x}_0$ is the sentence embedding of the instruction.

Each target token represents an imagination of the scene layout in its cell. In the initialization, these representations might be coarse and inaccurate, but our TD-STP refines the tokens progressively during navigation. At time step $t$, the target tokens of the previous step $\boldsymbol{c}_1^{t-1}, \dots, \boldsymbol{c}_q^{t-1}$ are sent into the transformer to update the token representation at time step $t$ with instruction and on-the-fly collected visual clues. The refined representations are then used to predict a more precise long-term target (navigation destination). Mathematically, at time step $t$, a multi-layer perceptron (MLP) and a softmax layer are used to obtain the probability of each target:
\begin{equation}
    \begin{aligned}\label{Eq:MLP}
    P_i^t = \text{softmax} \{ MLP(\boldsymbol{c}_i^t * \boldsymbol{g}^t)\},
    \end{aligned}
\end{equation}
where $\boldsymbol{g}^t$ is the global token, and $P_i^t$ indicates the likelihood of the navigation destination being closest to the $i$-th target (or equivalently, in the $i$-th cell) at time step $t$.

Equipping the agent with the ability to predict the long-term target leads to target-driven navigation. 
To better utilize this ability, we also add a positional embedding to the view tokens:
\begin{equation}
    \begin{aligned}
    \boldsymbol{v}_i^t \leftarrow \boldsymbol{v}_i^t + f_P(l_i^t), ~i\in \{1, 2, \dots, k^t\},
    \end{aligned}
\end{equation}
where $\boldsymbol{v}_i^t$ is the view feature of the $i$-th view at time step $t$, $f_P$ is the positional encoder in Eq.~\ref{eq:hist_emb}, and $l_i^t$ is the location that the $i$-th view refers to.
Therefore, all the tokens that represent actions in the action space contain a positional embedding, and thus the decision-making process can better utilize the guidance of the predicted long-term target location.

\definecolor{Gray}{gray}{0.94}
\begin{table*}
\centering
\tabcolsep=0.23cm
\caption{Comparison with state-of-the-art methods on the R2R dataset.}
\vspace{-2mm}
\label{tab:r2r}
\begin{tabular}{l>{\columncolor{Gray}}c>{\columncolor{Gray}}ccc|>{\columncolor{Gray}}r>{\columncolor{Gray}}ccc|>{\columncolor{Gray}}r>{\columncolor{Gray}}ccc}
\toprule
\multirow{2}{*}{Methods} & \multicolumn{4}{c|}{Validation Seen} & \multicolumn{4}{c|}{Validation Unseen} & \multicolumn{4}{c}{Test Unseen} \\
 & \multicolumn{1}{c}{SR$\uparrow$} & \multicolumn{1}{c}{SPL$\uparrow$} & \multicolumn{1}{c}{NE$\downarrow$} & \multicolumn{1}{c|}{OSR$\uparrow$} & \multicolumn{1}{c}{SR$\uparrow$} & \multicolumn{1}{c}{SPL$\uparrow$} & \multicolumn{1}{c}{NE$\downarrow$} & \multicolumn{1}{c|}{OSR$\uparrow$} & \multicolumn{1}{c}{SR$\uparrow$} & \multicolumn{1}{c}{SPL$\uparrow$} & \multicolumn{1}{c}{NE$\downarrow$} & \multicolumn{1}{c}{OSR$\uparrow$} \\ \hline\hline
Seq2Seq~\cite{anderson2018vln} & 39 & - & 6.01 & 53 & 22 & - & 7.81 & 28 & 20 & 18 & 7.85 & 27 \\
SF~\cite{fried2018speaker}& 66 & - & 3.36 & 74 & 35 & - & 6.62 & 45 & 35 & 28 & 6.62 & 44 \\
EnvDrop~\cite{tan2019learning} & 62 & 59 & 3.99 & - & 52 & 48 & 5.22 & - & 51 & 47 & 5.23 & 59 \\
OAAM~\cite{qi2020object} & 65 & 62 & - & 73 & 54 & 50 & - & 61 & 53 & 50 & - & 61 \\
AuxRN~\cite{zhu2020auxrn} & 70 & 67 & 3.33 & 78 & 55 & 50 & 5.28 & 62 & 55 & 51 & 5.15 & 62 \\
SERL~\cite{wang2020soft} & 69 & 64 & 3.20 & 75 & 56 & 48 & 4.74 & 65 & 53 & 49 & 5.63 & 61 \\
AP~\cite{wang2020active} & 70 & 52 & 3.20 & 80 & 58 & 40 & 4.36 & 70 & 60 & 41 & 4.33 & 71 \\
NvEM~\cite{an2021neighbor} & 69 & 65 & 3.44 & - & 60 & 55 & 4.27 & - & 58 & 54 & 4.37 & - \\
SSM~\cite{wang2021structured} & 71 & 62 & 3.10 & 80 & 62 & 45 & 4.32 & 73 & 61 & 46 & 4.57 & 70 \\
RecBERT~\cite{Hong2021vlnbert} & 72 & 68 & 2.90 & 79 & 63 & 57 & 3.93 & 69 & 63 & 57 & 4.09 & - \\
HAMT~\cite{chen2021hamt} & 76 & 72 & 2.51 & 82 & 66 & 61 & \textbf{2.29} & 73 & 65 & 60 & 3.93 & - \\ 
\hline
TD-STP (Ours) & \textbf{77} & \textbf{73} & \textbf{2.34} & \textbf{83} & \textbf{70} & \textbf{63} & 3.22 & \textbf{76} & \textbf{67} & \textbf{61} & \textbf{3.73} & \textbf{72}
\\
\bottomrule
\end{tabular}
\end{table*}

\begin{table*}
\centering
\tabcolsep=0.23cm
\caption{Comparison with state-of-the-art methods on the REVERIE dataset.}
\vspace{-2mm}
\label{tab:reverie}
\begin{tabular}{l>{\columncolor{Gray}}c>{\columncolor{Gray}}cccc|>{\columncolor{Gray}}c>{\columncolor{Gray}}cccc}
\toprule
\multirow{3}{*}{Methods}  & \multicolumn{5}{c|}{Validation Unseen} & \multicolumn{5}{c}{Test Unseen} \\
& \multicolumn{3}{c}{Navigation} & \multicolumn{2}{c|}{Grounding} & \multicolumn{3}{c}{Navigation} & \multicolumn{2}{c}{Grounding} \\
 & \multicolumn{1}{c}{SR$\uparrow$} & \multicolumn{1}{c}{SPL$\uparrow$} &  \multicolumn{1}{c}{OSR$\uparrow$}  & \multicolumn{1}{c}{RGS$\uparrow$} & \multicolumn{1}{c|}{RGSPL$\uparrow$} & \multicolumn{1}{c}{SR$\uparrow$} & \multicolumn{1}{c}{SPL$\uparrow$} & \multicolumn{1}{c}{OSR$\uparrow$} & \multicolumn{1}{c}{RGS$\uparrow$} & \multicolumn{1}{c}{RGSPL$\uparrow$} \\ \hline\hline
Seq2Seq~\cite{anderson2018vln} & 4.20 & 2.84 & 8.07 & 2.16 & 1.63 & 3.99 & 3.09 & 6.88 & 2.00 & 1.58 \\
RCM~\cite{wang2019reinforced} & 9.29 & 6.97 & 14.23 & 4.89 & 3.89 & 7.84 & 6.67 & 11.68 & 3.67 & 3.14 \\
SMNA~\cite{ma2019self} & 8.15 & 6.44 & 11.28 & 4.54 & 3.61 & 5.80 & 4.53 & 8.39 & 3.10 & 2.39 \\
FAST-MATTN~\cite{qi2020reverie} & 14.40 & 7.19 & 28.20 & 7.84 & 4.67 & 19.88 & 11.6 & 30.63 & 11.28 & 6.08 \\
SIA~\cite{lin2021scene}& 31.53 & 16.28 & \textbf{44.67} & \textbf{22.41} & 11.56 & 30.80 & 14.85 & \textbf{44.56} & 19.02 & 9.20 \\
RecBERT~\cite{Hong2021vlnbert} & 30.67 & 24.90 & 35.20 & 18.77 & 15.27 & 29.61 & 23.99 & 32.91 & 16.50 & 13.51 \\
HAMT~\cite{chen2021hamt}&  32.95 & \textbf{30.20} & 36.84 & 18.92 & \textbf{17.28} & 30.40 & 26.67 & 33.41 & 14.88 & 13.08 \\ \hline
TD-STP (Ours) &  \textbf{34.88} & 27.32 & 39.48 & 21.16 & 16.56 & \textbf{35.89} & \textbf{27.51} & 40.26 & \textbf{19.88} & \textbf{15.40} \\ \bottomrule
\end{tabular}
\end{table*}

\subsection{Model Optimization}
Following the common practice~\cite{chen2021hamt, moudgil2021soat, wang2021structured, Hong2021vlnbert, wang2019reinforced}, we adopt an imitation learning loss denoted as $\mathcal L_\text{IL}$ and a reinforcement learning loss denoted as $\mathcal L_\text{RL}$, and alternate between teacher forcing (using ground truth actions) and student forcing (using actions sampled from the policy). We further consider two extra losses: the former is to boost policy training in the global action space, while the latter is to supervise long-term target prediction.

The first loss function is used when the model is trained using sampled actions. Since the action space is expanded to include previously visited locations, the model is facing the problem that newly included actions are never chosen in teacher forcing. For better convergence during student forcing, we introduce \emph{history teacher loss}:
\begin{equation}
    \begin{aligned}
    \mathcal L_\text{HT} = -\sum\nolimits_{t=1}^T\log\pi(\bar a^t;\Theta),
    \end{aligned}
\end{equation}
where $\bar a^t$ is the action in the global action space that is on the ground truth trajectory and closest to the destination.

The second loss function relates to our IST mechanism. The target which is closest to the navigation destination is selected as the ground truth of target prediction. Thus a \emph{target prediction loss} can be derived:
\begin{equation}
    \begin{aligned}
    \mathcal L_\text T = -\sum\nolimits_{t=1}^T \log P_i^t,
    \end{aligned}
\end{equation}
where the $i$-th target token is closest to the navigation destination. The total loss function can be expressed as:
\begin{equation}
    \begin{aligned}
    \mathcal L = \alpha_1 \mathcal L_\text{IL} + \alpha_2 \mathcal L_\text{RL} + \alpha_3 \mathcal L_\text{HT} + \alpha_4 \mathcal L_\text T,
    \end{aligned}
\end{equation}
where $\alpha$s are coefficients.

\section{Experiment}
\subsection{Implementation Details}
We generally follow the transformer architecture and the corresponding hyper-parameters of~\cite{chen2021hamt}. The $MLP$s in Eq.~\ref{Eq:MLP1} and Eq.~\ref{Eq:MLP} are implemented by two linear layers with different weights. For IST, the grid size $d$ is set to $5$ forming a $5\times 5$ grid. The spacing between two adjacent positions $s$ is set to $6$ meters. In the ablation study, the two hyper-parameters are discussed in more detail.

As for model optimization, we alternate between teacher forcing and student forcing. In the teacher forcing iteration, the agent chooses the correct action and follows the ground-truth path. During this, reinforcement learning loss and history teacher loss are not used, and $\alpha_1$, $\alpha_2$, $\alpha_3$, $\alpha_4$ are set to $0.2$, $0$, $0$, $0.1$, respectively. In the student forcing iteration, the agent samples actions from the predicted probabilities. In this iteration, imitation learning loss is not used, and $\alpha_1$, $\alpha_2$, $\alpha_3$, $\alpha_4$ are set to $0$, $1$, $0.4$, $0.1$, respectively. More experimental results can be found at the supplementary material.

Our model is initialized with~\cite{chen2021hamt}, and trained on an NVIDIA V100 GPU for 100k iterations, with a batch size of $8$, a learning rate of 1e-5, and Adam optimizer~\cite{kingma2014adam}. 

\subsection{Performance on R2R Dataset}
\textbf{Dataset.}
R2R dataset~\cite{anderson2018vln} is based on Matterport3D Simulator~\cite{chang2017matterport3d} and consists of 90 houses with about 10k panoramic views. R2R has about 7k trajectories, and each trajectory has 3 instructions. The dataset is divided into 4 splits: the \emph{training split}, which consists of 61 houses and is used for training, the \emph{validation seen split}, which is used to validate the model in houses that are seen in the training split, the \emph{validation unseen split}, which consists of 11 houses that are not included in the previous two splits, and the \emph{test unseen split}, which consists of 18 houses that are not part of the previous 3 splits.
Among these splits, validation unseen and test unseen splits are relatively more important since they reflect the model's ability to generalize to previously unseen environments.

\vspace{0.5mm}
\noindent\textbf{Evaluation Metrics.}
We follow previous works in terms of evaluation metrics. Major evaluation metrics in R2R include the \emph{success rate} (SR), which is the ratio of navigating trajectories stopping 3 meters within the ground truth target, the \emph{success weighted by path length} (SPL), which is the success rate normalized by the ratio be-

\noindent tween the length of the ground-truth path and the agent's path, the \emph{navigation error}, which is the average distance between the agent's stopping point and the ground truth target, and the \emph{oracle success rate} (OSR), which is the success rate if the agent stops at the closest point to the destination in its trajectory.
Among these metrics, SR and SPL are relatively more important.

\vspace{0.5mm}
\noindent\textbf{Performance.}
The quantitative performance results are listed in Table~\ref{tab:r2r}. The results show that our proposed TD-STP achieves a consistent lead in terms of both SR and SPL. Noticeably, compared to the current SOTA~\cite{chen2021hamt}, our TD-STP achieves a larger improvement on SR and SPL in validation unseen and test unseen splits, which shows that our model better generalizes into unseen environments. Our model outperforms the current SOTA~\cite{chen2021hamt} in the validation unseen and the test unseen splits by \textbf{4\%} and \textbf{2\%} respectively in terms of 
SR. In addition to higher SR, our model also achieves higher SPL in unseen environments, which shows that our model can achieve a better trade-off between accuracy and efficiency.

\subsection{Performance on REVERIE Dataset}

\begin{table*}[h]
\caption{The ablated results of the main components on the R2R dataset.}
\vspace{-2mm}
\label{tab:main-ablation}
\center
\begin{tabular}{cc|c|c|c|cccc|>{\columncolor{Gray}}c>{\columncolor{Gray}}cccc}
\toprule
  & \multirow{2}{*}{Name} &   \multirow{2}{*}{ST} &\multirow{2}{*}{GAS}  & \multirow{2}{*}{IST} & \multicolumn{4}{c|}{Validation Seen} & \multicolumn{4}{c}{Validation Unseen}  \\
  &                       &                              &                              &                             & \multicolumn{1}{c}{SR$\uparrow$}   & \multicolumn{1}{c}{SPL$\uparrow$}  & \multicolumn{1}{c}{NE$\downarrow$}           & \multicolumn{1}{c|}{OSR$\uparrow$}  & \multicolumn{1}{c}{SR$\uparrow$}    & \multicolumn{1}{c}{SPL$\uparrow$}  & \multicolumn{1}{c}{NE$\downarrow$}     & \multicolumn{1}{c}{OSR$\uparrow$}    \\ \hline\hline
  &  Baseline       &                             &                              &                              & 75.0   & 71.7 &   2.51   &  81.9   & 65.7 & 60.9 &  3.65  & 73.4  \\     
    &   \#1                  & \checkmark                  &                    &
  & 73.7 & 70.9 &  2.71   &  79.3  & 67.7   & 62.5 &  3.50   &  74.8 \\
  &   \#2                  &   \checkmark                & \checkmark                   &
  & \textbf{77.1} & \textbf{73.0} &   2.40   &  82.0  &  68.5 & 62.4   &  3.32  & \textbf{76.5}  \\ 
  &   \#3                  & \checkmark                    &    \checkmark             & \checkmark 
  & 77.0 & 72.5  &   \textbf{2.34}   &  \textbf{82.9}  & \textbf{69.7}  & \textbf{62.7}   &  \textbf{3.22}  & 76.3 \\
\bottomrule
\end{tabular}
\end{table*}

\begin{table*}[h]
\caption{The ablated studies on the grid size and spacing in IST mechanism. Adopted parameters are marked with asterisks (*).}
\label{tab:ablation-design}
\vspace{-2mm}
\begin{subtable}[h]{0.45\linewidth}
\centering
\caption{The ablation study about the grid size ($d \times d$) in IST.}
\begin{tabular}{l|ccc|>{\columncolor{Gray}}c>{\columncolor{Gray}}cc} \toprule
\multirow{2}{*}{$d\times d$} & \multicolumn{3}{c|}{Validation Seen} & \multicolumn{3}{c}{Validation Unseen}  \\
 & \multicolumn{1}{c}{SR$\uparrow$} & \multicolumn{1}{c}{SPL$\uparrow$} & \multicolumn{1}{c|}{NE$\downarrow$}   & \multicolumn{1}{c}{SR$\uparrow$} & \multicolumn{1}{c}{SPL$\uparrow$} & \multicolumn{1}{c}{NE$\downarrow$}  \\  \hline\hline
$0\times 0$ & 77.1 & \textbf{73.0} & 2.40  & 68.5 & 62.4 & 3.32  \\ 
$3\times 3$ & \textbf{77.7} & \textbf{73.0} & 2.50  & 69.1 & 61.9 & 3.32  \\ 
$5\times 5$* &  77.0 & 72.5 & \textbf{2.34} & \textbf{69.7} & \textbf{62.7} & \textbf{3.22}    \\ 
$7\times 7$ &  76.2 & 70.5 & 2.49 & 68.9 & 60.8 & 3.27    \\ 
\bottomrule
\end{tabular}
\end{subtable}
\hspace{0.05\textwidth}
\begin{subtable}[h]{0.45\linewidth}
\caption{The ablated study about the spacing ($s$) in IST.}
\centering
\begin{tabular}{c|ccc|>{\columncolor{Gray}}c>{\columncolor{Gray}}cc} \toprule
\multirow{2}{*}{$s$ (meter)} & \multicolumn{3}{c|}{Validation Seen} & \multicolumn{3}{c}{Validation Unseen}  \\
 & \multicolumn{1}{c}{SR$\uparrow$} & \multicolumn{1}{c}{SPL$\uparrow$} & \multicolumn{1}{c|}{NE$\downarrow$}   & \multicolumn{1}{c}{SR$\uparrow$} & \multicolumn{1}{c}{SPL$\uparrow$} & \multicolumn{1}{c}{NE$\downarrow$}  \\  \hline\hline
4 &  77.4 & 71.1 & 2.32 &  68.9 & 61.6 & 3.36   \\ 
6* &  77.0 & 72.5 & 2.34  &  \textbf{69.7} & \textbf{62.7} & \textbf{3.22}  \\ 
8 &  \textbf{77.5} & \textbf{72.7} & \textbf{2.29} &  68.8 & 61.6 & 3.32  \\ 
10 &  76.1 & 70.8  & 2.41 &  68.3 & 61.7 & 3.28 \\ 
\bottomrule
\end{tabular}
\end{subtable}
\vspace{-1mm}
\end{table*}

\textbf{Dataset.}
Different from the R2R dataset, where the instructions are fine-grained, the REVERIE dataset~\cite{qi2020reverie} contains high-level instructions that ask the agent to find the described object. 
The REVERIE dataset has the same splits as the R2R dataset. 

\vspace{0.5mm}
\noindent\textbf{Evaluation Metrics.}
The evaluation metrics on REVERIE are simi-

\noindent lar to R2R. Major metrics include the \emph{success rate} (SR), which is the ratio of navigating trajectories stopping in places where the agent can see the target object, the \emph{success weighted by path length} (SPL), which is the success rate normalized by the ratio between the length of the ground-truth path and the agent's path, the \emph{oracle success rate} (OSR), which is the success rate if the agent stops at the closest point to the destination in its trajectory, the \emph{remote grounding success rate} (RGS), which is the success rate of finding the target object, and the \emph{remote grounding success weighted by path length} (RGSPL), which uses the ratio between the length of the ground-truth path and the agent's path to normalize RGS.

\vspace{0.5mm}
\noindent\textbf{Performance.}
Table~\ref{tab:reverie} shows the quantitative performance of our model on REVERIE compared to previous methods. Although our model achieves similar results compared to current SOTA methods in the validation unseen split (\emph{e.g.}, SIA~\cite{lin2021scene} and HAMT~\cite{chen2021hamt}), these methods suffer from overfitting when it comes to the test unseen split. Our model achieves a consistent lead in both navigation and grounding part of the dataset in the test unseen split. Most noticeably, our proposed model achieves a \textbf{16.5\%} relative improvement compared to the current SOTA in terms of SR (\emph{i.e.}, SIA). 

Since our model focuses mainly on navigation, we use a ViLBERT~\cite{lu2019vilbert} model fine-tuned on the REVERIE training set to perform the grounding task when navigation ends. 
Under this simple implementation, we still achieve new state-of-the-art performance in RGS and RGSPL in the test unseen split, showing the advantage of our model at high-level instructions.

\subsection{Ablation Studies}

\begin{figure*}[ht]
    \includegraphics[width=0.99\linewidth]{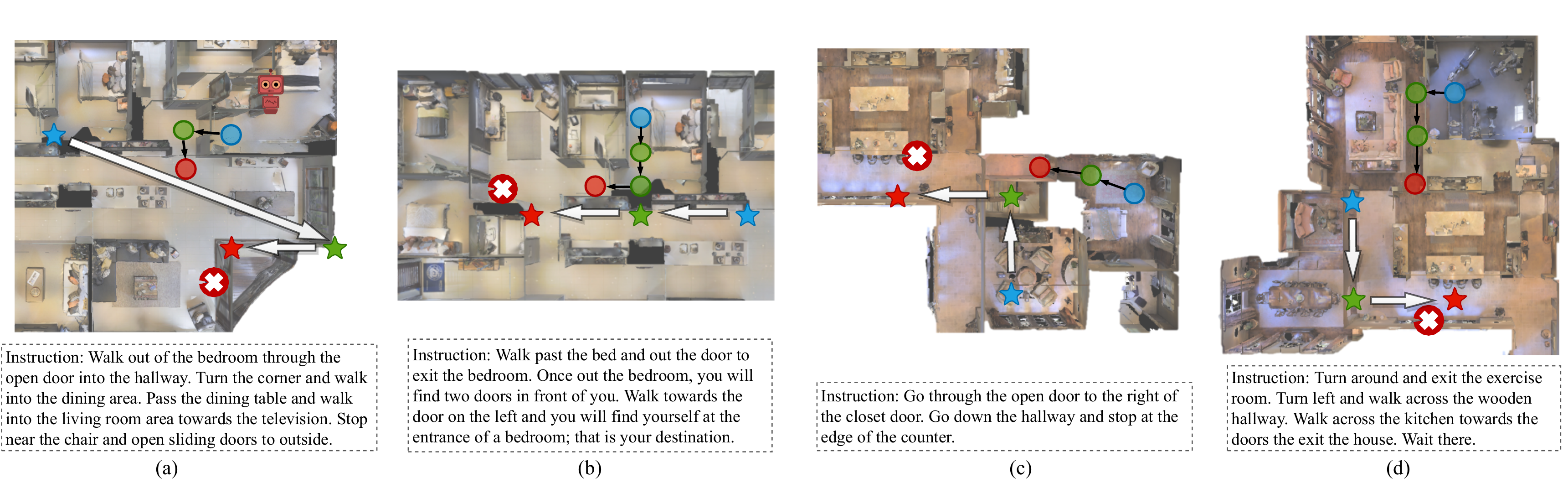}
    \vspace{-2.5mm}
    \caption{Visualization results. The circles represent navigation locations and the stars denote estimated navigation target. As the navigation proceeds (\protect\scalerel*{\includegraphics{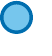}}{B}$\to$\protect\scalerel*{\includegraphics{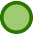}}{B}$\to$\protect\scalerel*{\includegraphics{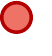}}{B}), the predicted targets are refined (\protect\scalerel*{\includegraphics{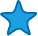}}{B}$\to$\protect\scalerel*{\includegraphics{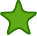}}{B}$\to$\protect\scalerel*{\includegraphics{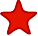}}{B}) and closer to the navigation destination (\protect\scalerel*{\includegraphics{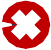}}{B}).}
    \label{fig:visualization}
    \vspace{-2mm}
\end{figure*}

\begin{figure}[t]
    \centering
    \includegraphics[width=0.85\linewidth]{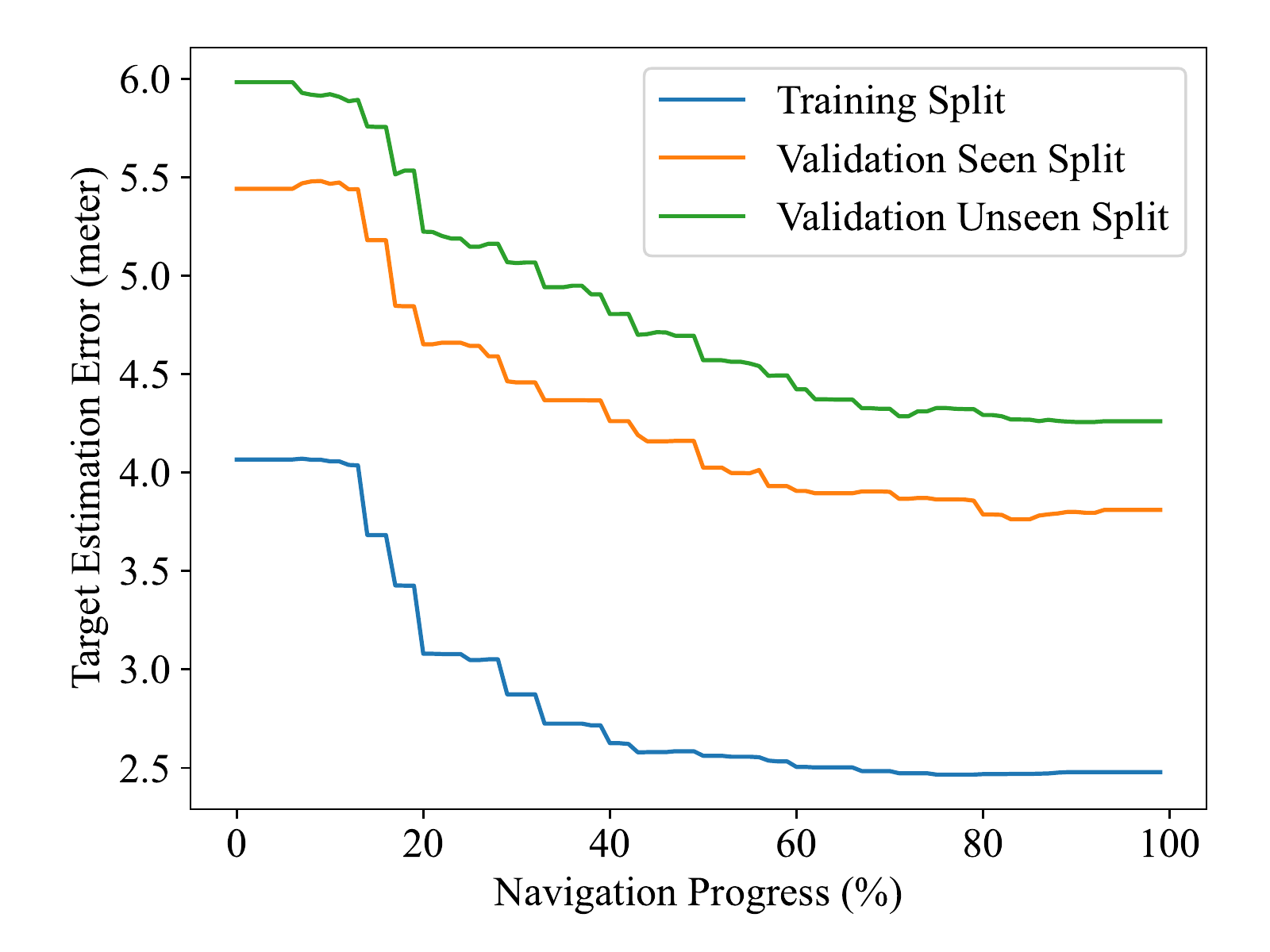}
        \vspace{-1mm}
    \caption{The trend of average target estimation error during inference in three splits. As the navigation progresses, the average target estimation error decreases.}
    \label{fig:tee}
    \vspace{-3mm}
\end{figure}

In this subsection, a set of ablation studies are conducted to verify the effectiveness of the proposed components, as shown in Table~\ref{tab:main-ablation}. Moreover, the design of the IST is also discussed in Table~\ref{tab:ablation-design}.

\vspace{0.5mm}
\noindent\textbf{Structured Transformer (ST).} In Table~\ref{tab:main-ablation}, compared with baseline~\cite{chen2021hamt}, "$\#1$" with the ST boosts SR and SPL from [$65.7\%$, $60.9\%$] to [$67.7\%$, $62.5\%$] respectively. It illustrates that adding the location information and topology relation of the history viewpoints into  the transformer benefits the navigation in unseen environments.

\vspace{0.5mm}
\noindent\textbf{Global Action Space (GAS).} 
As shown in Table~\ref{tab:main-ablation}, comparing "$\#2$" to "$\#1$", the GAS promotes SR from $73.7\%$ to $77.1\%$ in the validation seen set, and lifts SR from $67.7\%$ to $68.5\%$ in the validation unseen set. This demonstrates that flexibly jumping back to previously visited locations helps the agent to find the correct destination. The SPL in validation unseen split slightly drops because the backtracking of the GAS increases the trajectory length.

\vspace{0.5mm}
\noindent\textbf{Imaginary Scene Tokenization (IST) Mechanism.} In Table~\ref{tab:main-ablation}, comparing with "$\#2$", "$\#3$" with IST achieves another boost in both SR from $58.5\%$ to $69.7\%$ and SPL from $62.4\%$ to $62.7\%$ on validation unseen splits. It shows that the imagination of the target position is important for navigation in unknown environments.

\vspace{0.5mm}
\noindent\textbf{Grid Size of IST.} As shown in Table~\ref{tab:ablation-design} (a), we study the grid size of IST. Note that the grid size $0$ represents the model without IST. 
As can be seen in line 2, the model with a grid size of $3\times3$ achieves better navigation accuracy in both validation seen and unseen splits. When the grid size increases to $5\times5$, the model performs best on validation unseen split. 
However, when the grid size reaches $7\times7$, the SR and SPL on validation unseen split decrease, probably because too many target tokens may introduce noise into the model. We choose $5\times5$ as the setting in our final model.

\vspace{0.5mm}
\noindent\textbf{Spacing Size of IST.} As shown in Table~\ref{tab:ablation-design} (b), we study the spacing $s$ of two adjacent target locations, which controls the granularity of target candidate locations. The spacing of $6$ meters performs best in both SR and SPL of validation unseen split. An intuition for this is that a grid of targets that is too sparse provides little guidance for the agent, whereas a grid of targets that is too dense makes it hard for accurate target prediction.



\subsection{Analysis of Target-Driven Navigation}
\noindent\textbf{Qualitative Analysis.} 
As shown in Figure~\ref{fig:visualization}, we visualize how our TD-STP model modifies the estimated targets during navigation. At the beginning, the target estimation (blue star) is coarse, and when the agent takes one or two steps, the predicted targets (the stars in green and red) are closer to the destination. This illuminates our IST is able to refine its prediction with more information gathered. Therefore, the refined estimation provides better guidance for the decision-making process, boosting navigation performance.
\begin{figure}[t]
    \centering
    \includegraphics[width=0.88\linewidth]{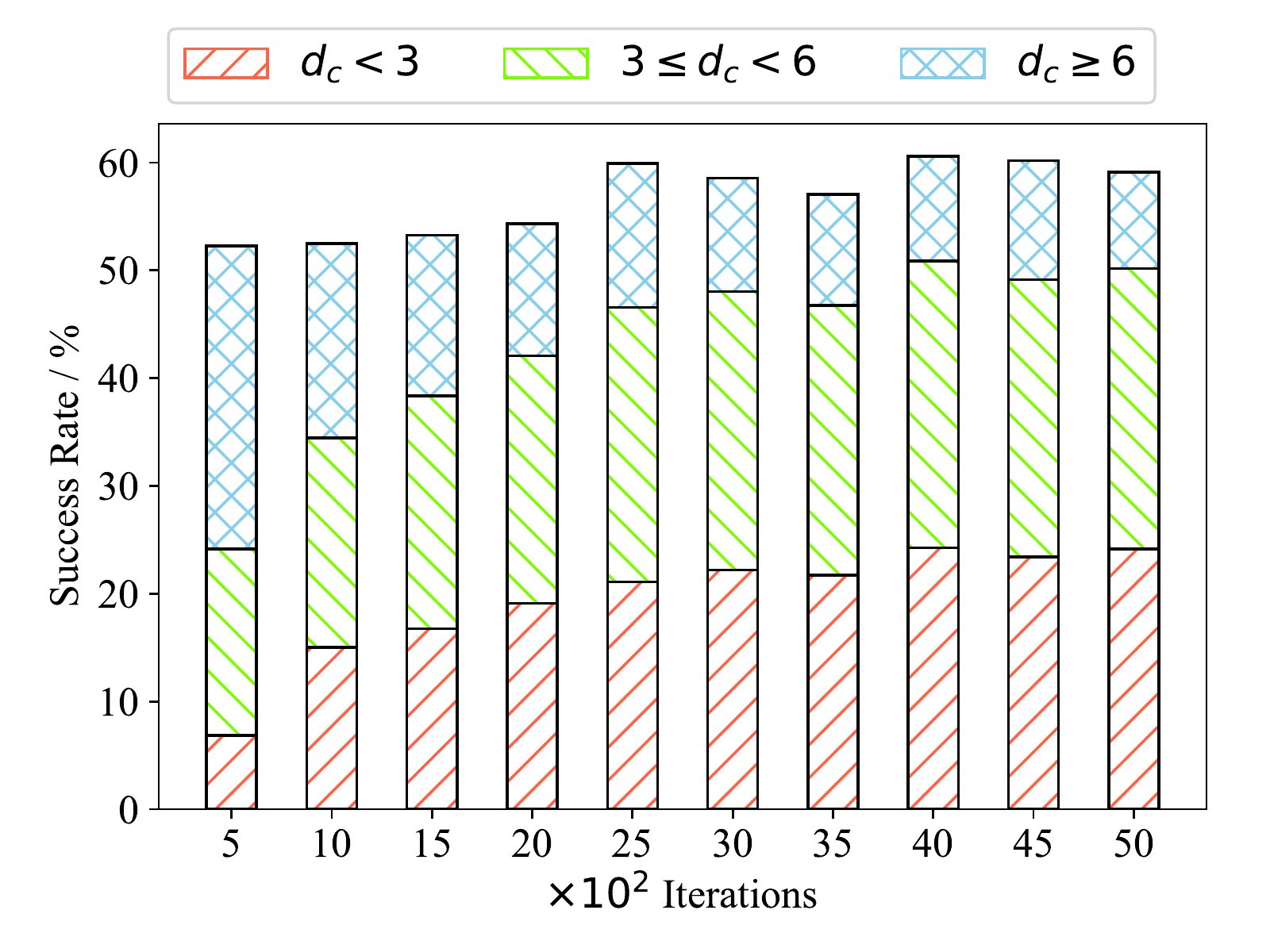}
    \caption{The SR on validation unseen split during training. Bars with different colors represent the target estimation error $d_c$ in different intervals. As training progresses, the proportion of ill-estimated targets ($d_c\ge 6$) decreases significantly, while the well-predicted targets ($d_c <3$) grows.}
    \label{fig:bar}
\end{figure}

\vspace{0.5mm}
\noindent\textbf{Quantitative Analysis.} 
We monitor the target estimation error at $50\%$ navigation process and the SR on validation unseen split during training, as shown in Figure~\ref{fig:bar}. The target estimation error $d_c$ is defined as the distance between the predicted target and the navigation destination. As the SR goes up, the proportion of ill-estimated targets ($d_c\ge 6$) drops substantially, while the well-estimated targets ($d_c<3$) increase significantly, which illuminates that the ability of target prediction gradually improves during training.

Figure~\ref{fig:tee} shows the average target prediction error as the navigation progresses during inference, which is conducted on three splits. As can be seen, the predicted target becomes closer to the navigation destination as the navigation progresses, which demonstrates the TD-STP can progressively refine the estimated target with more information collected during navigation.

\section{Conclusion}
In this paper, a Target-Driven Structured Transformer Planner (TD-

\noindent STP) is proposed for long-horizon goal-guided and room layout-aware navigation. TD-STP is built upon a Structured Transformer Planner (STP) with an Imaginary Scene Tokenization (IST) mechanism. Specifically, IST is for estimating the location of the final destination (typically located in the unexplored environment). By controlling information flow between input tokens (visited locations and estimated targets), STP achieves structured planning and global decision-making in an elegant and flexible manner. Extensive experiments demonstrate the superiority of our TD-STP. One limitation of this work is that TD-STP relies on the pre-defined environment graph. Thus a direction of our future effort is to incorporate SLAM technique for online map building.

{\noindent\small{\textbf{Acknowledgements}
This work was supported in part by the National Natural Science Foundation of China under Grant 62122010 and Grant 61876177, the Fundamental Research Funds for the Central Universities, the Key Research and Development Program of Zhejiang Province under Grant 2022C01082, and ARC DECRA DE220101390.}}

\newpage
\bibliographystyle{ACM-Reference-Format}
\balance 
\bibliography{sample-base}

\end{document}


\title{Supplementary Material: Target-Driven Structured Transformer Planner for Vision-Language Navigation}

\maketitle

\section{More Ablations}
In this section, additional ablation studies are provided, which study the weight of \emph{history teacher loss} and \emph{target prediction loss} (\emph{i.e.}$\alpha_3$ and $\alpha_4$). The experiments are conducted on the validation unseen split of R2R dataset~\cite{anderson2018vln} with Matterport3D Simulator~\cite{chang2017matterport3d}, and the results are listed in Table~\ref{tab:more-abl}.

\begin{table}[ht]
    \centering
    \caption{Ablation study about different values of $\alpha_3$ and $\alpha_4$. The adopted values are marked with asterisks.}
\begin{subtable}[h]{\linewidth}
\caption{The results in terms of SR}
\centering
\begin{tabular}{|ll|lll|}
\hline
\multicolumn{2}{|c|}{\multirow{2}{*}{SR $\uparrow$}}                & \multicolumn{3}{c|}{$\alpha_4$}                              \\ \cline{3-5} 
\multicolumn{2}{|c|}{}                                   & \multicolumn{1}{l|}{0.05} & \multicolumn{1}{l|}{0.1*} & 0.2  \\ \hline
\multicolumn{1}{|l|}{\multirow{3}{*}{$\alpha_3$}} & 0.2  & \multicolumn{1}{l|}{68.3} & \multicolumn{1}{l|}{68.4} & 68.4 \\ \cline{2-5} 
\multicolumn{1}{|l|}{}                            & 0.4* & \multicolumn{1}{l|}{69.3} & \multicolumn{1}{l|}{69.7} & 69.5 \\ \cline{2-5} 
\multicolumn{1}{|l|}{}                            & 0.6  & \multicolumn{1}{l|}{69.2} & \multicolumn{1}{l|}{69.6} & 68.5 \\ \hline
\end{tabular}
\end{subtable}

\begin{subtable}[h]{\linewidth}
\vspace{3mm}
\caption{The results in terms of SPL}
\centering
\begin{tabular}{|ll|lll|}
\hline
\multicolumn{2}{|c|}{\multirow{2}{*}{SPL $\uparrow$}}               & \multicolumn{3}{c|}{$\alpha_4$}                              \\ \cline{3-5} 
\multicolumn{2}{|c|}{}                                   & \multicolumn{1}{l|}{0.05} & \multicolumn{1}{l|}{0.1*} & 0.2  \\ \hline
\multicolumn{1}{|l|}{\multirow{3}{*}{$\alpha_3$}} & 0.2  & \multicolumn{1}{l|}{61.6} & \multicolumn{1}{l|}{62.5} & 62.3 \\ \cline{2-5} 
\multicolumn{1}{|l|}{}                            & 0.4* & \multicolumn{1}{l|}{62.2} & \multicolumn{1}{l|}{62.7} & 62.2 \\ \cline{2-5} 
\multicolumn{1}{|l|}{}                            & 0.6  & \multicolumn{1}{l|}{61.9} & \multicolumn{1}{l|}{61.9} & 61.4 \\ \hline
\end{tabular}
\end{subtable}
    \label{tab:more-abl}
\end{table}

As can be seen from the results, when $\alpha_3$ is set to $0.4$, the model achieves best performance with respect to SR and SPL. When $\alpha_3$ is too low, the backtracking process is not well-supervised, which hinders the global decision making. On the other hand, when $\alpha_3$ is too high, it breaks the balance between global decision making and other navigation processes, which results in lower SR and SPL. Similarly, when $\alpha_4$ is set to $0.1$, the model achieves the best performance. When $\alpha_4$ is too low, the target prediction process lacks proper supervision and therefore the performance is relatively undesirable compared to a higher $\alpha_4$. However, when the loss weight is too large, the model fails to achieve a balance between target prediction and current action selection, which leads to inferior performance.

\section{More Visualizations}
\begin{figure}[ht]
    \includegraphics[width=\linewidth]{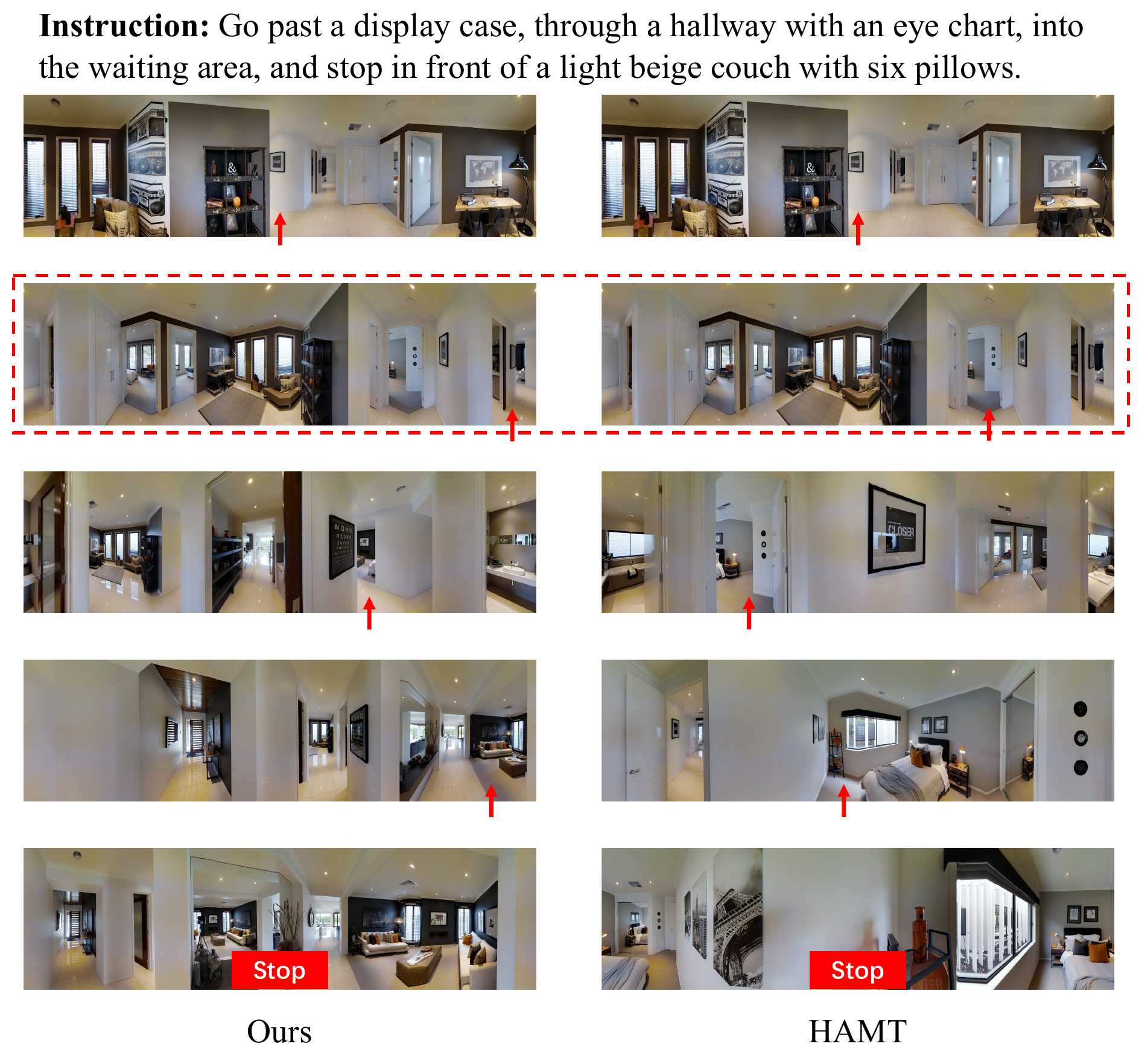}
    \caption{An example from the R2R validation unseen split. The panoramic views are displayed, and the red arrows denote the direction taken by the agents. We compare the navigation of our agent (left) and that of HAMT (right). The key difference between the two agents is the second step, which is highlighted. The result shows that our target-driven agent ends up in the right place and the HAMT agent ends up in the bedroom, which is far from the ground truth.}
    \label{fig:supp-vis-1}
    \vspace{1mm}
\end{figure}

\begin{figure}[ht]
    \includegraphics[width=\linewidth]{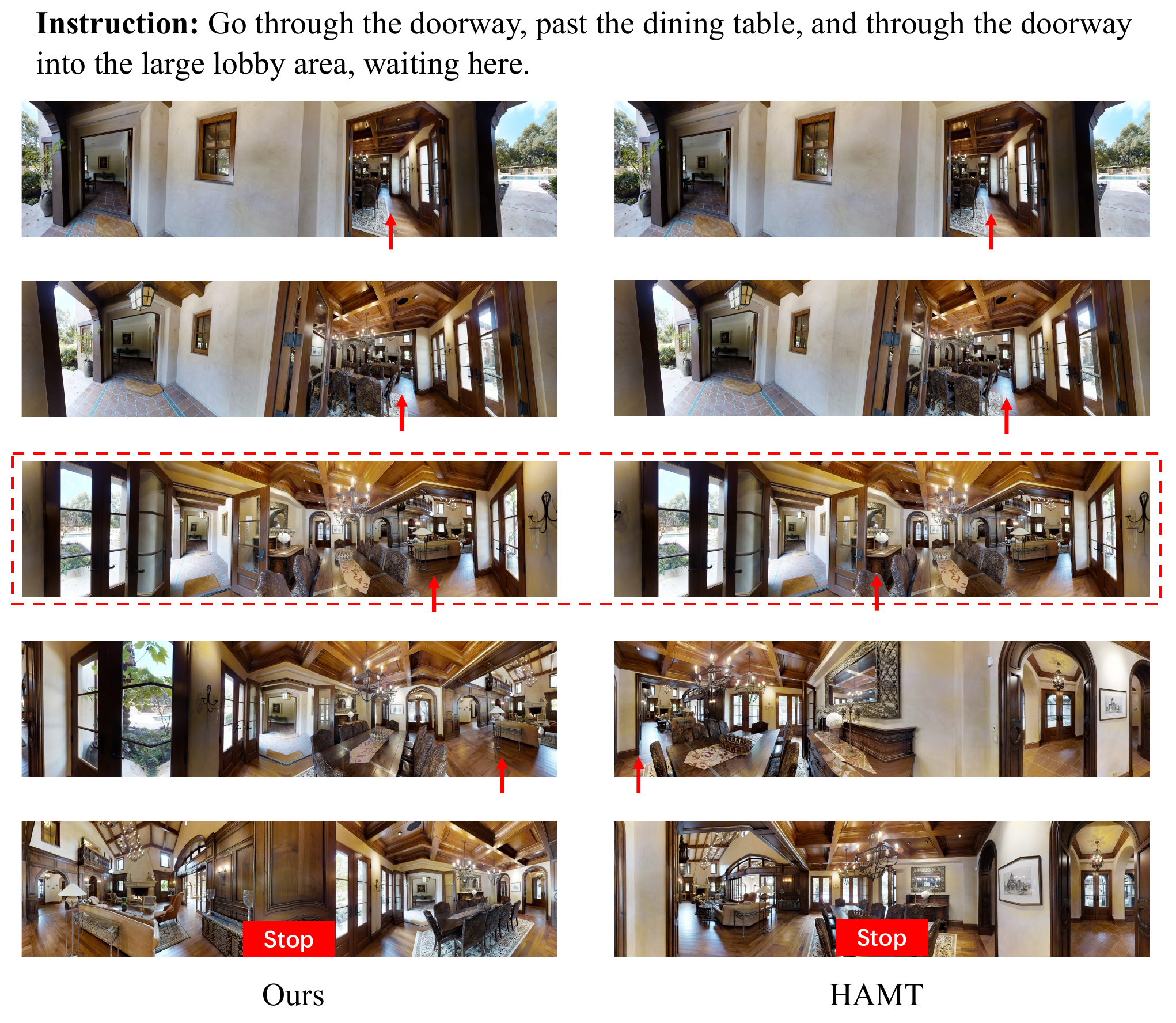}
    \caption{An example from the R2R validation unseen split. The comparison of TD-STP (ours) and HAMT offers another evidence of the importance of target-driven ability in navigation. In the third step, where the two agents differ, our agent predicts the navigation target (the lobby area) and heads to the direction of the target. By contrast, although the HAMT achieves good modeling of the history, it fails to look forward to the navigation future and walks in the wrong direction.}
    \label{fig:supp-vis-2}
    \vspace{1mm}
\end{figure}

In this section, additional visualization results are provided, as is shown in Figure~\ref{fig:supp-vis-1} and Figure~\ref{fig:supp-vis-2}. We compare the results of ours and those of HAMT~\cite{chen2021hamt} on the validation unseen split of R2R dataset~\cite{anderson2018vln}. The qualitative results demonstrate that the proposed TD-STP achieves better results with target-driven planning and structured modeling of the environment.

Figure~\ref{fig:supp-vis-1} compares our model with HAMT~\cite{chen2021hamt} on a challenging example. Note that the key difference is the second step. With the instruction "through the hallway", two possible directions can be observed: one taken by our TD-STP, the other taken by HAMT. Different from HAMT, which only considers the history information without modeling the future, our proposed TD-STP navigates with the guidance from the predicted target. Specifically, the proposed model is able to infer from the instruction and partially observed visual information the likely navigation destination, which is probably in the living room with a couch. In addition to statistical priors of typical room layouts, visual clues in the second step also provide some information. From the left hallway taken by HAMT, a nightstand can be vaguely seen, which offers a hint of a bedroom. By contrast, the correct direction that our agent selects leads to the edge of a couch, which is likely to be the navigation destination. The proposed TD-STP is able to infer from these visual-linguistic clues and estimate a likely target, which helps guide the navigation.

Figure~\ref{fig:supp-vis-2} offers another comparison of TD-STP and HAMT. The key difference is the third step, when the agent is supposed to "go past the dining table". Here, two possible ways of passing the dining table are available, and our agent selects the correct one which leads to the lobby area while the HAMT agent heads to the wrong direction, goes off the path, and loses direction. This example offers yet another evidence that being aware of the long-term target is crucial to navigation. At the third step, the lobby area is clearly in sight, but the HAMT agent fails to choose this direction. One explanation for this is that walking past the long edge of the dining table is statistically more common in the dataset and the agent adopts this prior. However, putting the navigation task aside, predicting the likely destination at the third step is relatively easy, and our agent is ready to utilize this predicted destination to guide navigation. Thus, the target-driven agent again outperforms the HAMT agent.


\newpage
\bibliographystyle{ACM-Reference-Format}
\balance 
\bibliography{sample-base}